%% file: acl2020.tex
\documentclass[11pt,a4paper]{article}
\usepackage[hyperref]{acl2020}
\usepackage{times}
\usepackage{enumerate}
\usepackage{verbatim}
\usepackage{latexsym}

\usepackage{microtype}

\aclfinalcopy 
\def\aclpaperid{2368} 

\usepackage{amsmath,amssymb,amsfonts}
\usepackage{url,enumitem}
\usepackage{booktabs}
\usepackage{xspace}
\usepackage{graphicx}
\usepackage{subcaption}
\usepackage[normalem]{ulem}
\newcommand{\Sref}[1]{\S\ref{#1}}

\title{Explaining Black Box Predictions and Unveiling Data Artifacts \\through Influence Functions}

\author{Xiaochuang Han \\
  Carnegie Mellon University \\
  {\small \tt xiaochuh@cs.cmu.edu} \\\And
  Byron C. Wallace \\
  Northeastern University \\
  {\small \tt b.wallace@northeastern.edu} \\\And
  Yulia Tsvetkov \\
  Carnegie Mellon University \\
  {\small \tt ytsvetko@cs.cmu.edu} \\}

\date{}

\begin{document}
\maketitle
\begin{abstract}
    \input{0_abstract}
\end{abstract}

\input{1_intro}
\input{2_background}

\input{3_experiment_setup.tex}
\input{4_rq1_rq2.tex}

\input{5_nli}
\input{6_related}
\input{7_conclusion}

\paragraph{Acknowledgments.}
We thank the anonymous ACL reviewers and members of TsvetShop at CMU for helpful discussions of this work.
This material is based upon work supported by NSF grants IIS1812327 and SES1926043, and by Amazon MLRA award.
Wallace's contributions were supported by the Army Research Office (W911NF1810328).
We also thank Amazon for providing GPU credits.

\bibliography{my_cites}
\bibliographystyle{acl_natbib}

\appendix
\input{A_implementation_details.tex}

\end{document}

%% file: 0_abstract.tex
Modern deep learning models for NLP are notoriously opaque. This has motivated the development of methods for interpreting such models, e.g., via gradient-based saliency maps or the visualization of attention weights. Such approaches aim to provide explanations for a particular model prediction by highlighting important words in the corresponding input text.
While this might be useful for tasks where decisions are explicitly influenced by individual tokens in the input, we suspect that such highlighting is not always suitable for tasks where model decisions should be driven by more complex reasoning. 
In this work, we investigate the use of \emph{influence functions} for NLP, providing an alternative approach to interpreting neural text classifiers.
Influence functions explain the decisions of a model by identifying influential training examples.
Despite the promise of this approach, influence functions have not yet been extensively evaluated in the context of NLP, a gap addressed by this work.
We conduct a comparison between influence functions and common word-saliency methods on representative tasks.
As suspected, we find that influence functions are particularly useful for natural language inference, a task in which `saliency maps' may not provide clear interpretation.
Furthermore, we develop a new quantitative measure based on influence functions that can reveal artifacts in training data.\footnote{Code is available at \url{https://github.com/xhan77/influence-function-analysis}.}

%% file: 1_intro.tex
\section{Introduction}

Deep learning models have become increasingly complex, and unfortunately their inscrutability has grown in tandem with their predictive power \cite{doshi2017towards}.
This has motivated efforts to design example-specific approaches to interpreting black box NLP model predictions, i.e., indicating specific input tokens as being particularly influential for a given prediction. This in turn facilitates the construction of \emph{saliency maps} over texts, in which words are highlighted with intensity proportional to continuous `importance' scores. Prominent examples of the latter include gradient-based attribution \citep{Simonyan2013DeepIC, Sundararajan2017AxiomaticAF, Smilkov2017SmoothGradRN}, LIME \citep{Ribeiro2016WhySI}, and attention-based \citep{Xu2015ShowAA} heatmaps.

While widely used and potentially useful for some lexicon-driven tasks (e.g., sentiment analysis), we argue that by virtue of being constrained to highlighting individual input tokens, saliency maps will necessarily fail to explain predictions in more complex semantic tasks involving reasoning, such as natural language inference (NLI), where fine-grained interactions between multiple words or spans are key \citep{Camburu2018eSNLINL}.
Moreover, saliency maps are inherently limited as a model debugging tool; they may tell us \emph{which} inputs the model found to be important, but not \emph{why}.

To address these shortcomings, we investigate the use of what \citet{Lipton2018TheMO} referred to as \emph{explanation by example}. Instead of constructing importance scores over the input texts on which the model makes predictions, such methods rank  \emph{training examples} by their influence on the model's prediction for the test input \citep{Caruana1999CasebasedEO, Koh2017UnderstandingBP, Card2019DeepWA}. Specifically, we are interested in the use of \emph{influence functions} \citep{Koh2017UnderstandingBP}, which are in a sense inherently `faithful'
in that they reveal the training examples most responsible for particular predictions. These do not require any modifications to the model structure.

\input{resources/first_fig.tex}

This paper presents a series of experiments intended to evaluate the potential utility of influence functions for better understanding modern neural NLP models. In this context, our contributions include answering the following research questions.

\begin{enumerate}[label={\bf \textsc{rq}\arabic*}]
\item We empirically assess whether the approximation to the influence functions \citep{Koh2017UnderstandingBP}
can be \emph{reliably} used to interpret decisions of deep transformer-based models such as BERT \cite{Devlin2019BERTPO}.

\item We investigate the degree to which results from the influence function are \emph{consistent} with insights gleaned from gradient-based saliency scores for representative NLP tasks.

\item We explore the application of influence functions as a mechanism to reveal \emph{artifacts} (or confounds) in training data that might be exploited by models.

\end{enumerate}

\noindent To the best of our knowledge, this is the first work in NLP to compare interpretation methods that construct saliency maps over inputs with methods that explain predictions via influential training examples. We also propose a new quantitative measurement for the effect of hypothesized artifacts \citep{Gururangan2018AnnotationAI, McCoy2019RightFT} on the model's prediction using influence functions.

%% file: resources/first_fig.tex
\begin{figure*}[t]
    \includegraphics[width=\textwidth]{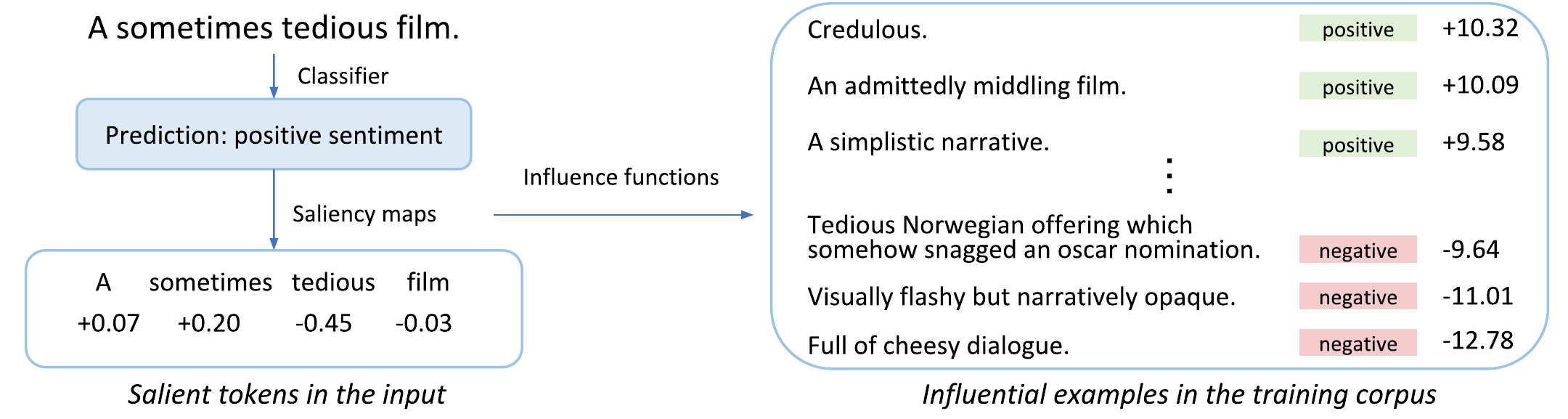}
    \caption{A sentiment analysis example interpreted by gradient-based saliency maps (left) and influence functions (right). Note that this example is classified incorrectly by the model. Positive saliency tokens and highly influential examples may suggest why the model makes the wrong decision; tokens and examples with negative saliency or influence scores may decrease the model's confidence in making that decision.
    }
    \label{fig:first_fig}
\end{figure*}

%% file: 2_background.tex
\section{Explaining Black-box Model Predictions}
\label{section:background}

Machine learning models in NLP depend on two factors when making predictions: the input text and the model parameters.
Prior attempts to interpret opaque NLP models have typically focused on the input text. Our work investigates the complementary approach of interpreting predictions by analyzing the influence of examples in training data.
Saliency maps aim to provide interpretability by highlighting parts of the input text, whereas influence functions seek clues in the model parameters, eventually locating interpretations within the training examples that influenced these estimates.
In this section we explain the two interpretation methods in detail.\footnote{
Here we focus on interpretability approaches which are faithful  \citep{Wiegreffe2019AttentionIN,Jacovi2020TowardsFI,jain2020} by construction; other approaches are discussed in \Sref{sec:related}.
}

\subsection{Gradient-based saliency maps}
As a standard, illustrative `explanation-by-input-features' method, we focus on gradient-based saliency maps, in which the gradient of the loss $\mathcal{L}$ is computed with respect to each token $t$ in the input text, and the magnitude of the gradient serves as a feature importance score \citep{Simonyan2013DeepIC,Li2016VisualizingAU}.
Gradients have the advantage of being locally `faithful' by construction: they tell us how much the loss would change, were we to perturb a token by a small amount.
Gradient-based attributions are also agnostic with respect to the model, as long as it is differentiable with respect to inputs. Finally, calculating gradients is computationally efficient, especially compared to methods that require post-hoc input perturbation and function fitting, like LIME \citep{Ribeiro2016WhySI}.

We are interested in why the model made a particular prediction. We therefore define a loss $\mathcal{L}_{\hat{y}}$ \emph{with respect to the prediction} $\hat{y}_i$ that the model actually made, rather than the ground truth $y_i$.
For each token $t \in x_i$, we define a saliency score $-\nabla_{e(t)}{\mathcal{L}_{\hat{y}}} \cdot e(t)$, where $e(t)$ is the embedding of $t$. This is also referred as the ``gradient $\times$ input'' method in \citet{Shrikumar2017LearningIF}. The ``gradient'' $\nabla_{e(t)}{\mathcal{L}_{\hat{y}}}$ captures the sensitivity of the loss to the change in the input embedding, and the ``input'' $e(t)$ leverages the sign and magnitude
of the input. The final saliency score of each token $t$ would be L1-normalized across all tokens in $x_i$.

Unlike \citet{Simonyan2013DeepIC} and \citet{Li2016VisualizingAU}, when scoring features for importance, we do not take the absolute value of the saliency score, as this encodes whether a token is positively influencing the prediction (i.e., providing support the prediction) or negatively influencing the prediction (highlighting counter-evidence). We show an example in the left part of \autoref{fig:first_fig}.

\subsection{Influence functions}
In contrast to explanations in the form of token-level heatmaps, the influence function provides a method for tracing model predictions back to training examples.
It first approximates how upweighting a particular training example $(x_i, y_i)$ in the training set $\{(x_1, y_1), \ldots, (x_n, y_n)\}$ by $\epsilon_i$ would change the learned model parameters $\hat{\theta}$:
\begin{align*}
    \frac{d\hat{\theta}}{d\epsilon_i} = - (\frac{1}{n} \sum_{j=1}^{n} \nabla_{\theta}^{2}{\mathcal{L}(x_j, y_j, \hat{\theta})})^{-1} \nabla_{\theta}{\mathcal{L}(x_i, y_i, \hat{\theta})}
\end{align*}
We can then use the chain rule to measure how this change in the model parameters would in turn affect the loss of the test input (as in saliency maps, w.r.t.~the model prediction):
\begin{align*}
    \frac{d\mathcal{L}_{\hat{y}}}{d\epsilon_i} = \nabla_{\theta}{\mathcal{L}_{\hat{y}}} \cdot \frac{d\hat{\theta}}{d\epsilon_i}
\end{align*}
More details (including proofs) can be found in \citet{Koh2017UnderstandingBP}.

We define the influence score for each training example $(x_i, y_i)$ as $-\frac{d\mathcal{L}_{\hat{y}}}{d\epsilon_i}$, and then $z$-normalize it across all examples in the training set. Note that since $\mathcal{L}_{\hat{y}}$ is defined with respect to a particular test input, influence scores of training examples are also defined for individual test instances.

Intuitively, a \emph{positive} influence score for a training example means: were we to remove this example from the train set, we would expect a \emph{drop} in the model's confidence when making the prediction on the test input. A \emph{negative} influence score means that removing the training example would increase the model's confidence in this prediction. We show an example in the right part of \autoref{fig:first_fig}.

%% file: 3_experiment_setup.tex
\section{Experimental Setup}
\label{sec:setup}
We are interested in analyzing and comparing the two interpretation approaches (gradient-based attributions and influence functions) on relatively shallow, lexicon-driven tasks and on more complex, reasoning-driven tasks. We focus on sentiment analysis and natural language inference (NLI) as illustrative examples of these properties, respectively. Both models are implemented on top of BERT encoders \citep{Devlin2019BERTPO}. In particular we use BERT-Base, with the first 8 of the 12 layers frozen, only fine-tuning the last 4 transformer layers and the final projection layer.\footnote{We used smaller BERT models because influence functions are notoriously expensive to compute. We also resort to the same stochastic estimation method, LiSSA \citep{Agarwal2016SecondOrderSO}, as in \citet{Koh2017UnderstandingBP}, and we deliberately reduce the size of our training sets. Even with these efforts, computing the influence scores of 10k training examples w.r.t. one typical test input would take approximately 10 minutes on one NVIDIA GeForce RTX 2080 Ti GPU. }

It is worth noting that influence functions are guaranteed to be accurate only when the model is strictly convex (i.e., its Hessian is positive definite and thus invertible) and is trained to convergence. However, deep neural models like BERT are not convex, and one often performs early stopping during training. We refer to \citet{Koh2017UnderstandingBP} for details on how influence functions can nonetheless provide good approximations. To summarize briefly: for the non-convexity issue, we add an appropriate `damping' term to the model's Hessian so that it is positive definite and invertible. Concerning non-convergence: the approximated influence may still be interpretable as the true influence of each training example plus a constant offset that does not depend on the individual examples. Aside from this theory, we also perform a sanity check in \Sref{sec:rq1} to show that influence functions can be applied to BERT in practice on the two tasks that we consider.

\paragraph{Sentiment analysis}
We use a binarized version of the \textit{Stanford Sentiment Treebank (SST-2)} \citep{Socher2013RecursiveDM}. Our BERT-based model is trained on 10k examples; this achieves 89.6\% accuracy on the SST-2 dev set of 872 examples.
We randomly sample 50 examples from the SST-2 dev set as the set for which we extract explanations for model predictions.

\paragraph{Natural language inference}
Our deeper `semantic' task is NLI, a classification problem that concerns the relationship between a premise sentence and a hypothesis sentence. NLI is a ternary task with three types of premise--hypothesis relations: \emph{entailment}, \emph{neutral}, and \emph{contradiction}.
We train our BERT model on the Multi-Genre NLI (MNLI) dataset \citep{Williams2017ABC}, which contains 393k premise and hypothesis pairs of three relations from 10 different genres. We collapse the \emph{neutral} and \emph{contradiction} labels to a single \emph{non-entailment} label and only use 10k randomly sampled examples for training. On the MNLI dev set of 9815 examples, the model achieves an accuracy of 84.6\%.

To evaluate model interpretations in a controlled manner, we adopt a diagnostic dataset, HANS \citep{McCoy2019RightFT}. This contains a balanced number of examples where hypotheses may or may not entail premises with certain artifacts that they call `heuristics' (e.g., lexical overlap, subsequence). The original HANS dataset contains 30k examples that span 30 different heuristic sub-categories. We test our model and interpretation methods on 30 examples covering all the sub-categories.

%% file: 4_rq1_rq2.tex
\section{Evaluating Influence Functions for NLP}
\label{sec:rq1}

\paragraph{\textsc{rq}1: Is influence function approximation reliable when used for deep architectures in NLP?}
Influence functions are designed to be an approximation to leave-one-out training for each training example.
But the theory only proves that this works on strictly convex models. While \citet{Koh2017UnderstandingBP} show that influence functions can be a good approximation even when the convexity assumption is not satisfied (in their case, a CNN for image classification), it is still not obvious that the influence function would work for BERT.

Therefore, we conduct a sanity check: for each instance in our test set, we by turns remove the most \emph{positively} influential 10\%, the most \emph{negatively} influential 10\%, the \emph{least} influential (where influence scores are near zero) 10\%, and a \emph{random} 10\% of training examples. We are interested in how these removals in retraining would affect the confidence of model predictions. \autoref{tab:SA_sanity_check} and \autoref{tab:NLI_sanity_check} show the result of experiments on sentiment analysis and NLI, repeated with 5 random initialization seeds.

\input{resources/SA_sanity_check.tex}
\input{resources/NLI_sanity_check.tex}

The results are largely in accordance with our expectation in both tasks: removing the most positively influential training examples would cause the model to have a significantly lower prediction confidence for each test example; removing the most negatively influential examples makes the model slightly more confident during prediction; and removing the least influential examples leads to an effect that is closest to removing a same amount of random examples (although we note that deleting the least influential features still yields a larger $\Delta$ than choosing features at random to remove in NLI). We therefore conclude that the influence function behaves reasonably and reliably for BERT in both sentiment analysis and NLI tasks.

\paragraph{\textsc{rq}2. Are gradient-based saliency maps and `influential' examples compatible?
}
Comparing saliency maps and outputs from application of the influence function is not straightforward. Saliency maps communicate the importance of individual tokens in test instances, while influence functions measure the importance of training examples. Still, it is reasonable to ask if they seem to tell similar stories regarding specific predictions. We propose two experiments that aim to estimate the \emph{consistency} between these two interpretation methods.

The first experiment addresses \emph{whether a token with high saliency also appears more frequently in the training examples that have relatively high influence}. For each example in the test set, we find the tokens with the most positive, most negative, and median saliency scores. We then find all the influential training examples w.r.t.~the test inputs that contain one of these tokens. These training examples could have any labels in the label set. We further only consider examples whose label is the same as the test prediction, because the token saliency scores, whether positive or negative, are directly w.r.t.~the test prediction, and the effect of a token in an oppositely labeled training example is therefore indirect.

We compute the average influence score of these training examples and report the results on top 10\%, 20\%, 50\%, and all training examples for both sentiment analysis and NLI tasks in \autoref{fig:SA_direct_consistency} and \autoref{fig:NLI_direct_consistency} respectively. The reason we have results at different granularity is that from empirical results in \citet{Koh2017UnderstandingBP}, we see that the influence function approximations tend to be less accurate when going from the most influential to the less influential examples down in the spectrum.

\input{resources/SA_fig_direct_consistency.tex}
\input{resources/NLI_fig_direct_consistency.tex}

In the task of sentiment analysis, we observe that training examples containing the most positively salient token in the test example generally have a higher influence to the test prediction.
However, we do not see this trend (in fact, it is the opposite) in the task of natural language inference.

The second experiment answers the question of \emph{whether the influence result would change significantly when a salient token is removed from the input}. Again, for each of the test examples, we identify the tokens with the most positive, most negative, and median saliency score. We by turns remove them from the input and compute the influence distribution over all training examples. We compare these new influence results with the one on the original input, and report an overlap rate of the top 0.1\%, 0.2\%, 0.5\%, and 1\% influential training examples before and after the token removal. \autoref{tab:SA_indirect_consistency} and \autoref{tab:NLI_indirect_consistency} show results for sentiment analysis and NLI, respectively.

\input{resources/SA_indirect_consistency.tex}
\input{resources/NLI_indirect_consistency.tex}

When removing a token with the most positive saliency score, we expect the model to be less confident about its current prediction; it could possibly make a different prediction. Therefore, we expect to see a most different influence distribution from the original influence result compared to removing the token with median or the most negative saliency score. This is exactly what we observe in \autoref{tab:SA_indirect_consistency} for sentiment analysis. However, for NLI, we again see a rather opposite trend: removing the most negatively salient token (might make the prediction more confident but should not change the prediction itself) leads to the most different influence distribution.

We conclude from the above two experiments that gradient-based saliency maps and influential examples are compatible and consistent with each other in sentiment analysis. However, for NLI the two approaches do not agree with each other and could potentially tell very different stories. To this end, we take a closer look at the task of NLI.

%% file: resources/SA_sanity_check.tex
\begin{table}[h]
\small
\begin{center}
\renewcommand{\arraystretch}{1.3}
\begin{tabular}{p{0.17\textwidth}r}
    \toprule
    {Removal type} & Avg. $\Delta$ in prediction confidence\\
    \midrule
    {Positively influential} & $-6.00\%$ ($\pm 1.12\%$)\\
    {Negative influential} & $+0.17\%$ ($\pm 0.50\%$)\\
    {Least influential} & $-1.30\%$ ($\pm 0.54\%$)\\
    {Random} & $-1.67\%$ ($\pm 0.54\%$)\\
    \bottomrule
\end{tabular} 
\end{center}
\caption{
Sanity check for influence function result on BERT in sentiment analysis.
}
\label{tab:SA_sanity_check}
\end{table}

%% file: resources/NLI_sanity_check.tex
\begin{table}[h]
\small
\begin{center}
\renewcommand{\arraystretch}{1.3}
\begin{tabular}{p{0.17\textwidth}r}
    \toprule
    {Removal type} & Avg. $\Delta$ in prediction confidence\\
    \midrule
    {Positively influential} & $-11.62\%$ ($\pm 2.09\%$)\\
    {Negative influential} & $+2.01\%$ ($\pm 1.44\%$)\\
    {Least influential} & $+1.01\%$ ($\pm 0.97\%$)\\
    {Random} & $+0.13\%$ ($\pm 1.07\%$)\\
    \bottomrule
\end{tabular} 
\end{center}
\caption{
Sanity check for influence function result on BERT in NLI.
}
\label{tab:NLI_sanity_check}
\end{table}

%% file: resources/SA_fig_direct_consistency.tex
\begin{figure}[h]
    \includegraphics[width=\linewidth]{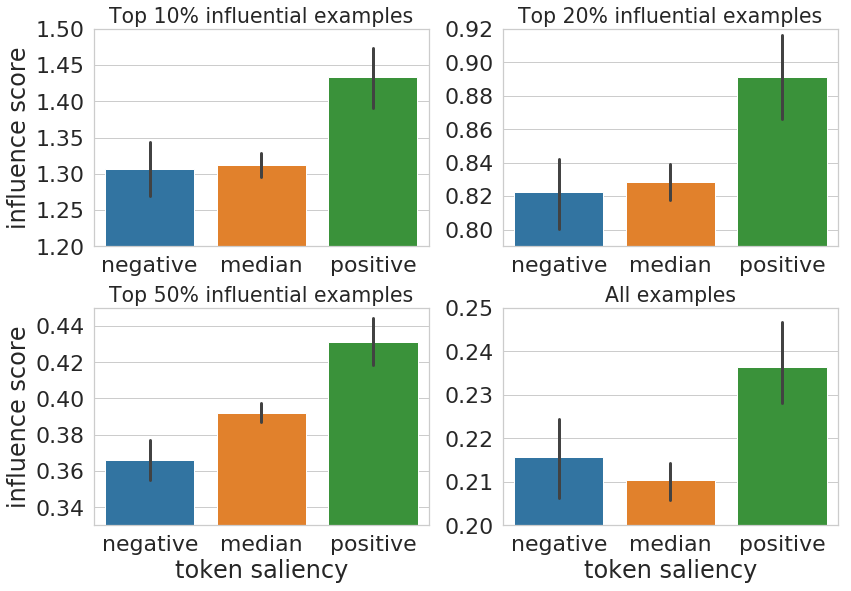}
    \caption{Average \textbf{influence score} of top \textit{sentiment analysis} training examples that contain a token in test example with most positive, most negative, or median saliency. Error bars depict standard errors.
    }
    \label{fig:SA_direct_consistency}
\end{figure}

%% file: resources/NLI_fig_direct_consistency.tex
\begin{figure}[h]
    \includegraphics[width=\linewidth]{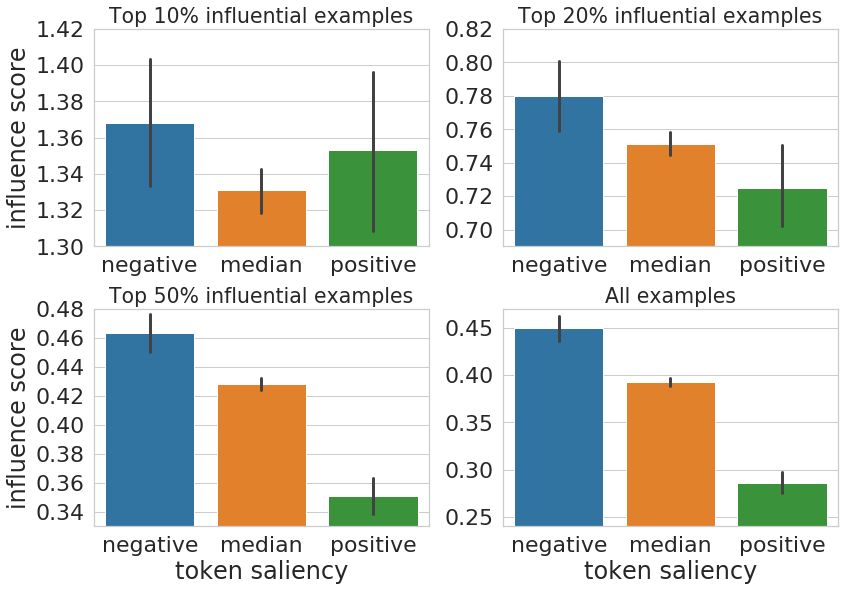}
    \caption{Average \textbf{influence score} of top \textit{NLI} training examples that contain a token in test example with most positive, most negative, or median saliency. Standard error is shown in error bars.}
    \label{fig:NLI_direct_consistency}
\end{figure}

%% file: resources/SA_indirect_consistency.tex
\begin{table}[t]
\small
\begin{center}
\renewcommand{\arraystretch}{1.3}
\begin{tabular}{p{0.12\textwidth}rrrr}
    \toprule
    {Saliency of the removed token} & @0.1\% & @0.2\% & @0.5\% & @1\%\\
    \midrule
    {Most negative} & 75.6\% & 77.4\% & 80.0\% & 82.4\%\\
    {Median} & 84.2\% & 86.7\% & 88.9\% & 89.1\%\\
    {Most positive} & 65.2\% & 68.8\% & 71.4\% & 72.0\%\\
    \bottomrule
\end{tabular} 
\end{center}
\caption{
Average \textbf{overlap rate} of top influential \textit{sentiment analysis} training examples before and after removal of a token with the most positive, most negative, or median saliency.
}
\label{tab:SA_indirect_consistency}
\end{table}

%% file: resources/NLI_indirect_consistency.tex
\begin{table}[t]
\small
\begin{center}
\renewcommand{\arraystretch}{1.3}
\begin{tabular}{p{0.12\textwidth}rrrr}
    \toprule
    {Saliency of the removed token} & @0.1\% & @0.2\% & @0.5\% & @1\%\\
    \midrule
    {Most negative} & 33.0\% & 33.5\% & 37.5\% & 40.9\%\\
    {Median} & 79.3\% & 78.0\% & 80.5\% & 84.0\%\\
    {Most positive} & 46.0\% & 48.3\% & 49.9\% & 54.9\%\\
    \bottomrule
\end{tabular} 
\end{center}
\caption{
Average \textbf{overlap rate} of top influential \textit{NLI} training examples before and after removal of a token with the most positive, negative, or median saliency.
}
\label{tab:NLI_indirect_consistency}
\end{table}

%% file: 5_nli.tex
\section{Interpreting NLI Predictions with Influence Functions}

\paragraph{Are saliency-based explanations useful for NLI?}
Gradient-based saliency maps are faithful by construction, but this does not mean that they will highlight input tokens that humans find plausible or useful. We hypothesize that highlighting individual input tokens as important is likely most useful for `shallow' classification tasks like sentiment analysis, and less so for more complex reasoning tasks such as NLI.

To contrast the types of explanations these methods offer in this context, we show explanations for a prediction made for a typical example in HANS in the form of a saliency map and influential examples in \autoref{tab:NLI_qual_example_1}. The tokens that get the most positive and most negative saliency scores are marked in cyan and red, respectively. The training examples with the most positive and most negative influence scores are presented as supporting and opposing instances, respectively.

\input{resources/NLI_example.tex}

The relationship classification decision in NLI is often made through an interaction between multiple words or spans. Therefore, an importance measure on each individual token might not give us much useful insight into model prediction. Though influence functions also do not explicitly tell us which latent interactions between words or spans informed the model prediction, we can test whether the model is relying on some hypothesized artifacts in a post-hoc way by looking at patterns in the influential training examples.

In \autoref{tab:NLI_qual_example_1}, though the most influential examples (both supporting and opposing) are ostensibly far from the test input, they all exhibit lexical overlap between the premise and hypothesis. Some of the influential training examples (e.g., the 4th supporting example and 2nd opposing example) capture a reverse ordering of spans in the premise and hypothesis. We note that our test input also has a high lexical overlap and similar reverse ordering. This exposes a problem: the model might be relying on the wrong artifacts like word overlap during the decision process rather than learning the relationship between the active and passive voice in our case. This problem was surfaced by finding influential examples.

\subsection{Quantitatively measuring artifacts}
\citet{McCoy2019RightFT} hypothesize that the main artifact NLI models might learn is \emph{lexical overlap}. In fact, for all of the examples in HANS, every word in the hypothesis would appear in the corresponding premise (100\% lexical overlap rate).
Half of the examples would have an \emph{entailment} relationship while the other half have an \emph{non-entailment} relationship. \citet{McCoy2019RightFT} compare four models with strong performance in MNLI, and all of them predict far more entailments than non-entailments.
Because of this imbalance in prediction, they conclude that the models are perhaps exploiting artifacts in data when making decisions.

We see one potential problem out of the above method: it can only be applied to a certain group of examples and imply a \emph{general} model behavior by examining the prediction imbalance. However, model behavior should depend on the actual example it sees each time. The extent to which the model exploits the artifact in each \emph{individual} example remains unclear.

To analyze the effect of artifacts on individual examples, we propose a method using influence functions. We hypothesize that if an artifact informs the model's predictions for a test instance, the most influential training examples for this test example should contain occurrences of said artifact. For instance, if our model exploits `lexical overlap' when predicting the relation between a premise and a hypothesis, we should expect the most influential training examples found by the influence function to have a highly overlapping premise and hypothesis.

In \autoref{fig:1-1}, we plot each training example's influence score and lexical overlap rate between its premise and hypothesis for a typical example in the HANS dataset. In linen with our expectation, the most influential (both positively and negatively) training examples tend to have a higher lexical overlap rate. Note that we also expect this trend for the most negatively influential examples, because they influence the model's prediction as much as the positively influential examples do, only in a different direction.

To quantify this bi-polarizing effect, we find it natural to fit a quadratic regression to the influence-artifact distribution. We would expect a high positive quadratic coefficient if the artifact feature appears more in the most influential examples. For an irrelevant feature, we would expect this coefficient to be zero. With this new quantitative measure, we are ready to explore the below problems unanswered by the original diagnostic dataset.

\paragraph{For test examples predicted as non-entailment, did the model fail to recognize that they have a lexical overlap feature? Was the artifact
not exploited in these cases?}
\autoref{fig:1-1} and \autoref{fig:1-2} show two examples in HANS, one predicted as entailment and the other predicted as non-entailment. We observe that the example predicted as non-entailment does not have a significantly different influence-artifact pattern from the entailment example.
In fact, the average quadratic coefficients for all examples predicted as entailment and non-entailment are $+3.28 \times 10^{-3}$ and $+3.30 \times 10^{-3}$ respectively.
Therefore, for predicted non-entailment examples, we still see that the most influential training examples tend to have a high rate of lexical overlap, indicating that the model still recognizes the artifact in these cases.

\input{resources/NLI_quadratic_figure_1.tex}
\input{resources/NLI_quadratic_figure_2.tex}

\paragraph{The model relies on training examples with high lexical overlap when predicting in the artificial HANS dataset. Would it still exploit the same artifact for natural examples?}
Apart from finding the most influential training examples for each HANS example, we also apply influence functions on 50 natural MNLI examples, not controlled to exhibit any specific artifacts. A typical example is shown in \autoref{fig:1-3}.
The average quadratic coefficient over all 50 natural examples is $+0.65 \times 10^{-3}$, which is considerably smaller than the above cases in HANS dataset. The model therefore does not rely on as much lexical overlap in natural examples as in the diagnostic dataset.

\paragraph{We have been analyzing scenarios focusing on one data artifact. What if we have a second artifact during prediction possibly indicating a contradicting decision? How will the model recognize the two artifacts in such a scenario?}
We know that lexical overlap could be a data artifact exploited by NLI models for making an \emph{entailment} prediction in HANS.
On the other hand, as briefly pointed out by \citet{McCoy2019RightFT}, other artifacts like negation might be indicative of \emph{non-entailment}.
We are interested in how two contradicting artifacts might compete when they both appear in an example.
We take all examples in HANS labeled as entailment and manually negate the hypothesis so that the relationship becomes non-entailment.
For example, a hypothesis ``the lawyers saw the professor'' would become ``the lawyers did not see the professor''.

\autoref{fig:2-1} and \autoref{fig:2-2} show the influence-artifact distributions on both lexical overlap and negation for an original HANS example.
\autoref{fig:2-3} and \autoref{fig:2-4} show the distributions for the same HANS example with negated hypothesis.
The average quadratic coefficients on all examples are shown in \autoref{tab:NLI_coefficient_1}.
We observe that in the original HANS example, negation is actually a \emph{negative} artifact: the training examples with negation tend to be the least influential ones. In the negated HANS example, we see the effect of negations becomes positive, while the effect of lexical overlap is drastically weakened. This confirms that the model recognizes the new set of artifacts, and the two are competing with each other.

\input{resources/NLI_coefficient_1.tex}

Importantly, observing an artifact in the most influential training examples is a \emph{necessary} but not \emph{sufficient} condition to concluding that it was truly exploited by the model. However, it can serve as a first step towards identifying artifacts in black-box neural models and may be complemented by probing a larger set of hypothesized artifacts.

%% file: resources/NLI_example.tex
\begin{table}[h]
\small
\begin{center}
\renewcommand{\arraystretch}{1.4}
\begin{tabular}{p{0.33\textwidth}r}
    \toprule
    \textbf{Test input}\\
    {\textit{P:} The \uwave{\textcolor{red}{manager}} was \uwave{\textcolor{red}{encouraged}} by the secretary. \textit{H:} The secretary \uline{\textcolor{cyan}{encouraged}} \uline{\textcolor{cyan}{the}} manager.} & \{entail\}\\
    \midrule
    \textbf{Most \uline{\textcolor{cyan}{supporting}} training examples}\\
    {\textit{P:} Because you're having fun. \textit{H:} Because you're having fun.} & [entail]\\
    {\textit{P:} I don't know if I was in heaven or hell, said Lillian Carter, the president's mother, after a visit. \textit{H:} The president's mother visited.} & [entail]\\
    {\textit{P:} Inverse price caps. \textit{H:} Inward caps on price.} & [entail]\\
    {\textit{P:} Do it now, think 'bout it later. \textit{H:} Don't think about it now, just do it.} & [entail]\\
    \midrule
    \textbf{Most \uwave{\textcolor{red}{opposing}} training examples}\\
    {\textit{P:} H'm, yes, that might be, said John. \textit{H:} Yes, that might be the case, said John.} & [non-entail]\\
    {\textit{P:} This coalition of public and private entities undertakes initiatives aimed at raising public awareness about personal finance and retirement planning. \textit{H:} Personal finance and retirement planning are initiatives aimed at raising public awareness.} & [non-entail]\\
    \bottomrule
\end{tabular} 
\end{center}
\caption{
A correctly predicted example in HANS interpreted by saliency map and influence function.
}
\label{tab:NLI_qual_example_1}
\end{table}

%% file: resources/NLI_quadratic_figure_1.tex
\begin{figure*}[ht]
    \centering
    \begin{subfigure}[h]{0.32\textwidth}
        \includegraphics[width=\textwidth]{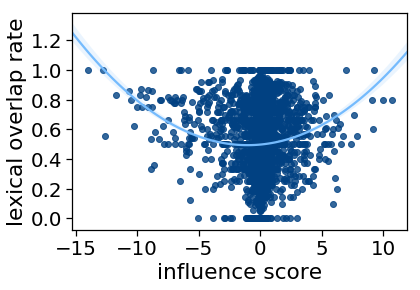}
        \caption{HANS example predicted as \textbf{entailment}. (\textit{P:} The athlete by the doctors encouraged the senator. \textit{H:} The athlete encouraged the senator.) Quadratic coefficient: $+3.74 \times 10^{-3}$.}
        \label{fig:1-1}
    \end{subfigure}
    \hfill
    \begin{subfigure}[h]{0.32\textwidth}
        \includegraphics[width=\textwidth]{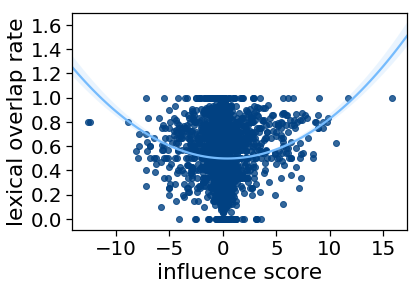}
        \caption{HANS example predicted as \textbf{non-entailment}. (\textit{P:} Since the author introduced the actors, the senators called the tourists. \textit{H:} The senators called the tourists.) Quadratic coef: $+3.59 \times 10^{-3}$.}
        \label{fig:1-2}
    \end{subfigure}
    \hfill
    \begin{subfigure}[h]{0.32\textwidth}
        \includegraphics[width=\textwidth]{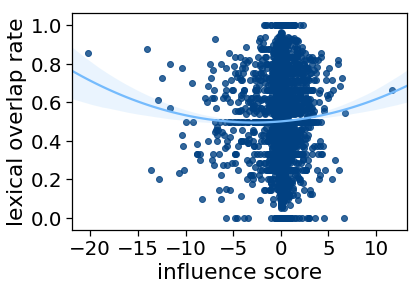}
        \caption{A typical MNLI example. (\textit{P:} And uh as a matter of fact he's a draft dodger. \textit{H:} They dodged the draft, I'll have you know.) Quadratic coefficient: $+0.74 \times 10^{-3}$.}
        \label{fig:1-3}
    \end{subfigure}
    
    \caption{Influence-artifact distribution for different test examples.}
\end{figure*}

%% file: resources/NLI_quadratic_figure_2.tex
\begin{figure*}[ht]
    \centering
    \begin{subfigure}[h]{0.23\textwidth}
        \includegraphics[width=\textwidth]{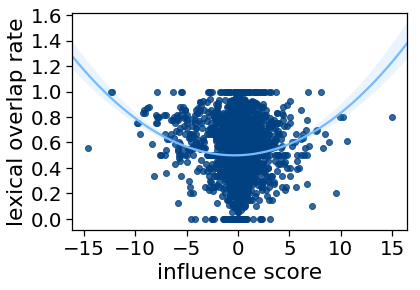}
        \caption{Lexical overlap in original HANS example. Quadratic coefficient: $+3.13 \times 10^{-3}$.}
        \label{fig:2-1}
    \end{subfigure}
    \hfill
    \begin{subfigure}[h]{0.23\textwidth}
        \includegraphics[width=\textwidth]{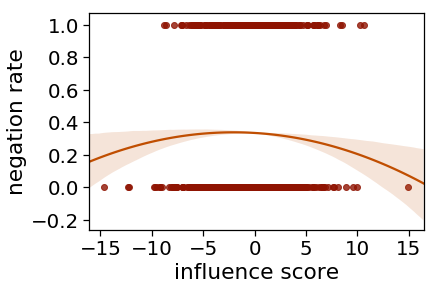}
        \caption{Negation in original HANS example. Quadratic coefficient: $-0.92 \times 10^{-3}$.}
        \label{fig:2-2}
    \end{subfigure}
    \hfill
    \begin{subfigure}[h]{0.23\textwidth}
        \includegraphics[width=\textwidth]{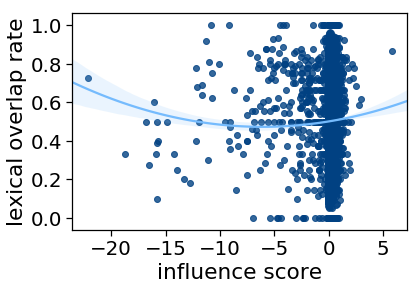}
        \caption{Lexical overlap in negated HANS example. Quadratic coefficient: $+0.76 \times 10^{-3}$.}
        \label{fig:2-3}
    \end{subfigure}
    \hfill
    \begin{subfigure}[h]{0.23\textwidth}
        \includegraphics[width=\textwidth]{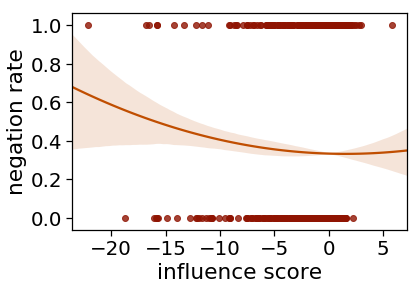}
        \caption{Negation in negated HANS example. Quadratic coefficient: $+0.55 \times 10^{-3}$.}
        \label{fig:2-4}
    \end{subfigure}
    
    \caption{Influence-artifact distribution for an original and negated HANS example. (\textit{P:} The lawyers saw the professor behind the bankers. \textit{H:} The lawyers \textbf{saw} / \textbf{did not see} the professor.)}
\end{figure*}

%% file: resources/NLI_coefficient_1.tex
\begin{table}[t]
    \centering
    \begin{tabular}{@{}lrr@{}}
    \toprule
     & Lexical overlap coef & Negation coef\\
      \midrule
      Original & $+3.05 \times 10^{-3}$ & $-1.13 \times 10^{-3}$\\
      Negated & $+0.53 \times 10^{-3}$ & $+0.27 \times 10^{-3}$\\
      \bottomrule
    \end{tabular}
    \caption{Average \textbf{quadratic coefficients} of the influence-artifact distribution for all original HANS examples and all negated HANS examples.}
    \label{tab:NLI_coefficient_1}
\end{table}

%% file: 6_related.tex
\section{Related Work}
\label{sec:related}
Interpreting NLP model predictions by constructing importance scores over the input tokens is a widely adopted approach \cite{belinkov2019analysis}. Since the appearance and rise of attention-based models, many work naturally inspect attention scores and interpret with them. However, we are aware of the recent discussion over whether attention is a kind of faithful explanation \citep{Jain2019AttentionIN, Wiegreffe2019AttentionIN}. Using vanilla attention as interpretation could be more problematic in now ubiquitous deep transformer-based models, such as we use here. 

Gradient-based saliency maps are locally `faithful' by construction. Other than the vanilla gradients \citep{Simonyan2013DeepIC} and the ``gradient $\times$ input'' method \citep{Shrikumar2017LearningIF} we use in this work, there are some variants that aim to make gradient-based attributions robust to potential noise in the input \citep{Sundararajan2017AxiomaticAF, Smilkov2017SmoothGradRN}. We also note that \citet{Feng2018PathologiesON} find that gradient-based methods sometimes yield counter-intuitive results when iterative input reductions are performed.

Other token-level interpretations include input perturbation \citep{Li2016UnderstandingNN} which measure a token's importance by the effect of removing it, and LIME \citep{Ribeiro2016WhySI} which can explain any model's decision by fitting a sparse linear model to the local region of the input example.

The main focus of this work is the applicability of influence functions \citep{Koh2017UnderstandingBP} as an interpretation method in NLP tasks, and to highlight the possibility of using this to surface annotation artifacts. Other methods that can trace the model's decision back into the training examples include deep weighted averaging classifiers \citep{Card2019DeepWA}, which make decisions based on the labels of training examples that are most similar to the test input by some distance metrics. \citet{Croce2019AuditingDL} use kernel-based deep architectures that project test inputs to a space determined by a group of sampled training examples and make explanations through the most activated training instances. While these methods can similarly identify the `influential' training examples, they require special designs or modifications to the model and could sacrifice the model's performance and generalizability.

Other general methods for model interpretability include adversarial-attack approaches that identify that part of input texts can lead to drastically different model decisions when minimally edited \citep{Ebrahimi2017HotFlipWA, Ribeiro2018SemanticallyEA}, probing approaches that test internal representations of models for certain tasks and properties \citep{Liu2019LinguisticKA, Hewitt2019DesigningAI}, and generative approaches that make the model jointly extract or generate natural language explanations to support predictions \citep{Lei2016RationalizingNP, Camburu2018eSNLINL, Liu2018TowardsEN, rajani2019explain}.

Specific to the NLI task, \citet{Gururangan2018AnnotationAI} recognize and define some possible artifacts within NLI annotations. \citet{McCoy2019RightFT} create a diagnostic dataset that we use in this work and suggest that the model could be exploiting some artifacts in training data based on its poor performance on the diagnostic set. Beyond NLI, the negative influence of artifacts in data was explored in other text classification tasks \citep{Pryzant2018DeconfoundedLI,kumar-etal-2019-topics,landeiro2019discovering}, focusing on approaches to adversarial learning to demote the artifacts.

%% file: 7_conclusion.tex
\section{Conclusion}
We compared two complementary interpretation methods---gradient-based saliency maps and influence functions---in two text classification tasks: sentiment analysis and NLI. We first validated the reliability of influence functions when used with deep transformer-based models. We found that in a lexicon-driven sentiment analysis task, saliency maps and influence functions are largely consistent with each other. They are not consistent, however, on the task of NLI. We posit that influence functions may be a more suitable approach to interpreting models for such relatively complex natural language `understanding` tasks (while simpler attribution methods like gradients may be sufficient for tasks like sentiment analysis).

We introduced a new potential use of influence functions: revealing and quantifying the effect of data artifacts on model predictions, which have been shown to be very common in NLI. Future work might explore how rankings induced over training instances by influence functions can be systematically analyzed in a stand-alone manner (rather than in comparison with interpretations from other methods), and how these might be used to improve model performance. Finally, we are interested in exploring how these types of explanations are actually interpreted by users, and whether providing them actually establishes trust in predictive systems.

%% file: A_implementation_details.tex
\section{Implementation Details}
The main model we used for experiments is a BERT-Base model \citep{Devlin2019BERTPO}, adapted from \citet{Wolf2019HuggingFacesTS}. We ``froze'' the embedding layer and the first 8 transformer layers and only fine-tuned the last 4 transformer layers and the final projection layer. We used the default BERT optimizer with default hyperparameters: a learning rate of $5\mathrm{e}{-5}$, a total of $3$ epochs, a max sequence length of $128$, and a training batch size of $32$.

For gradient-based saliency maps, we used a ``vanilla'' version implemented by \citet{Wallace2019AllenNLPIA}. For influence functions, we adapted code from \citet{Koh2017UnderstandingBP} to PyTorch and used the same stochastic estimation trick, LiSSA \citep{Agarwal2016SecondOrderSO}. Since our model is not convex, we used a ``damping'' term (as mentioned in \Sref{sec:setup}) of $3\mathrm{e}{-3}$. This value was picked so that the recursive approximation to the inverse Hessian-vector product can be finished (converged) in a reasonable time. More specifically, we chose the recursion depth to be $2500$ (with a total of 10k training examples), the number of recursions to be $1$, and a scaling factor to be $1\mathrm{e}{4}$. In each step estimating the Hessian-vector product, we took a batch of $8$ training examples for stability. We empirically checked that the inverse Hessian-vector product converges after the recursive estimation for all test examples on which we performed the analysis.